\pdfoutput=1

\documentclass[11pt]{article}

\usepackage[final]{acl}

\usepackage{times}
\usepackage{latexsym}

\usepackage[T1]{fontenc}

\usepackage[utf8]{inputenc}
\usepackage{microtype}
\usepackage{inconsolata}
\usepackage{graphicx}
\usepackage{booktabs} 
\usepackage{colortbl}
\usepackage{multirow}
\usepackage{amsmath}
\usepackage{listings} 
\usepackage{xcolor}
\usepackage{authblk} 

\title{Multimodal Inconsistency Reasoning (MMIR): \\ A New Benchmark for Multimodal Reasoning Models}

\author[1]{Qianqi Yan}
\author[1]{Yue Fan}
\author[ ]{Hongquan Li}
\author[2]{Shan Jiang}
\author[2]{Yang Zhao}
\author[2]{Xinze Guan}
\author[2]{\\Ching-Chen Kuo}
\author[1]{Xin Eric Wang}
\affil[1]{University of California, Santa Cruz}
\affil[2]{eBay}

\begin{document}

\maketitle

\begin{abstract}
     Existing Multimodal Large Language Models (MLLMs) are predominantly trained and tested on consistent visual-textual inputs, leaving open the question of whether they can handle inconsistencies in real-world, layout-rich content. To bridge this gap, we propose the Multimodal Inconsistency Reasoning (MMIR) benchmark to assess MLLMs' ability to detect and reason about semantic mismatches in artifacts such as webpages, presentation slides, and posters. MMIR comprises 534 challenging samples, each containing synthetically injected errors across five reasoning-heavy categories: Factual Contradiction, Identity Misattribution, Contextual Mismatch, Quantitative Discrepancy, and Temporal/Spatial Incoherence.
     We evaluate eight state-of-the-art MLLMs, showing that models with dedicated multimodal reasoning capabilities, such as o1, substantially outperform their counterparts while open-source models remain particularly vulnerable to inconsistency errors.
     Detailed error analyses further show that models excel in detecting pairwise inconsistencies but struggle with inconsistencies confined to single elements in complex layouts. 
     Probing experiments reveal that single-modality prompting, including Chain-of-Thought (CoT) and Set-of-Mark (SoM) methods, yields marginal gains, revealing a key bottleneck in cross-modal reasoning. Our findings highlight the need for advanced multimodal reasoning and point to future research on multimodal inconsistency.

\end{abstract}

\section{Introduction}

\begin{figure}[h!]
\setlength{\abovecaptionskip}{0.1cm}
    \centering
    \includegraphics[width=\columnwidth]{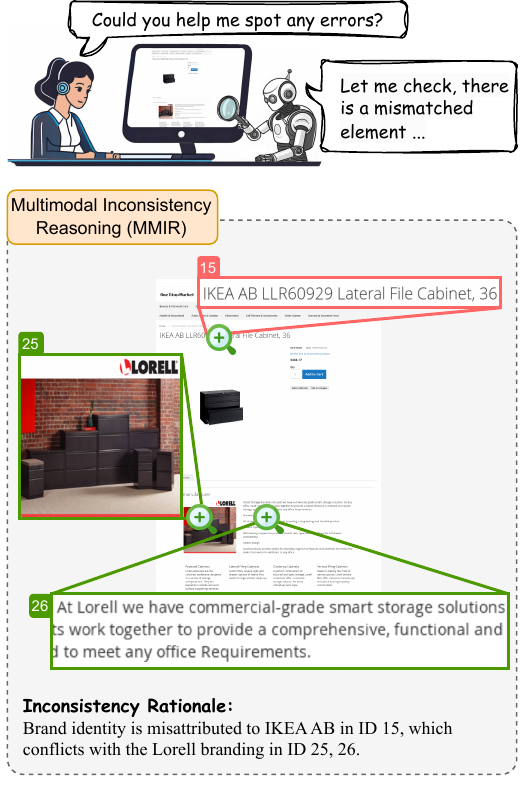}
    \caption{\textbf{An illustration of multimodal inconsistency reasoning on a webpage.} An agent examines a webpage where the brand “IKEA AB” is mentioned, but other elements clearly refer to “Lorell.” Detecting this brand identity misattribution requires the ability to compare text fields across different sections of the page and reconcile them with accompanying images or context—an inherently multimodal reasoning task.
    }
    \label{fig:teaser}
\end{figure}

Recent advances in Large Language Models (LLMs) have demonstrated impressive reasoning abilities across a variety of tasks~\cite{openai2024gpto1card, guo2025deepseek, kojima2022large, chain-of-thought}. Building on pre-trained LLMs, Multimodal Large Language Models (MLLMs) are evolving rapidly. However, they usually face greater challenges as they need to reason across different modalities, especially when inconsistencies (i.e., mismatched or contradictory contents) exist. We find that, being primarily trained and evaluated on consistent visual-textual inputs, existing MLLMs are largely untested in scenarios where the input contains misaligned or contradictory information---a situation that is common in real-world scenarios. For example, in Figure~\ref{fig:teaser}, a user presents a web page containing conflicting visual and textual elements, asking the model to identify errors.

To comprehensively evaluate the ability of MLLMs in reasoning over multimodal inconsistency, we introduce the \textbf{Multimodal Inconsistency Reasoning Benchmark (MMIR)}. MMIR is the first framework dedicated to evaluating how effectively MLLMs can reason about and identify semantic mismatches within complex, layout-rich content with interleaved image and text components. 
Our benchmark is built on a diverse collection of real-world artifacts (e.g., websites, slides, posters) which have been augmented with \textbf{synthetic inconsistencies}---realistic inconsistency errors injected into their original structures. These inconsistency errors span a range of reasoning-heavy categories: \textit{Factual Contradiction, Identity Misattribution, Contextual Mismatch, Quantitative Discrepancy}, and \textit{Temporal/Spatial Incoherence}, posing a next-level reasoning challenge for models. 
For example, resolving a \textit{Identity Misattribution} involves verifying entity alignment across modalities, while \textit{Quantitative Discrepancy} requires cross-referencing chart data with textual claims. 
By challenging models to detect such inconsistencies, MMIR forces them to perform intricate reasoning that goes well beyond simple pattern recognition. 
Our benchmark not only exposes the limitations of current MLLMs in handling real-world challenges of reasoning over multimodal content with inconsistency, but also provides a platform for developing more robust multimodal reasoning systems.

In our experiments, we evaluated the advanced multimodal reasoning model o1~\cite{openai2024gpto1card} and seven other state-of-the-art MLLMs: GPT-4o~\cite{openai2024gpt4ocard}, Qwen2.5-VL~\cite{Qwen2.5-VL}, Llama-3.2~\cite{grattafiori2024llama}, LLaVA-NeXT~\cite{liu2024llavanext}, InternVL2.5~\cite{Chen2024ExpandingPB}, and Phi-3.5-Vision~\cite{abdin2024phi3technicalreporthighly} using MMIR's 534 test samples. 
The results overall underscore that current MLLM models struggle with multimodal inconsistency reasoning. Specifically, there is a stark contrast between proprietary and open-source models. The best open-source model evaluated only reaches less than 40\% accuracy. o1 with strong reasoning capability achieves the overall best performance with over 50\% accuracy.

To further understand the benchmarking results, we conduct an analysis based on the inconsistency category, modality, and layout complexity of the artifact. We find the proprietary models excel in identifying factual contradiction and identity misattribute types of inconsistency and pairwise inconsistency, either inter-modality or intra-modality.
Last but not least, we investigate some approaches to enhance the model's performance in our probing experiment. The results indicate that text-based Chain-of-Thought prompting and visual-based prompting (Set-of-Mark annotations) offer minimal and sometimes adverse effects, whereas an iterative multimodal interleaved reasoning strategy shows promising gains. Overall, these results highlight a critical bottleneck in the ability of MLLMs to perform robust, integrated reasoning—a key challenge for future research.

Our contributions are threefold: 
\begin{itemize} 
\item We introduce MMIR, a novel benchmark that targets the critical yet underexplored task of multimodal inconsistency reasoning in layout-rich content. 
\item We perform a comprehensive evaluation of one leading multimodal reasoning model and seven state-of-the-art MLLMs, revealing significant gaps in their ability to detect inconsistency errors with detailed error analyses across multiple error types, modalities, and layout complexities.
\item We provide detailed probing analyses that expose key challenges---from perceptual shortcomings to reasoning bottlenecks---and propose a framework that iteratively refines predictions by jointly leveraging visual and textual modalities.
\end{itemize}

\section{Related Work}

\paragraph{Multimodal Understanding and Reasoning} 
Multimodal Large Language Models (MLLMs) process multimodal inputs by first processing visual inputs with pre-trained vision encoders such as CLIP \cite{radford2021learning} to extract features, and then projecting them into the textual representation space with adapters \cite{liu2024improved,li2023blip}. Significant efforts have been made to bridge the gap between vision and text modalities via integrating more cross-modality data such as interleaved image-text sequences and visual grounding data \cite{alayrac2022flamingo,chen2023shikra, peng2023kosmos}. Also, some recent works develop MLLMs with improved nuanced multimodal abilities, such as Optical Character Recognition (OCR) \cite{bai2023qwen,liu2024llavanext}, layout understanding \cite{feng2024layoutgpt, fan2024read}, Graphic User Interface (GUI) interpretation \cite{liu2024harnessingwebpageuistextrich, Qwen2.5-VL}.

As MLLMs typically leverage pre-trained large language models (LLMs) as the backbone, they inherent strong textural reasoning abilities from the advanced LLMs\cite{floridi2020gpt,touvron2023llama, bai2023qwen,taori2023stanford,chowdhery2023palm, openai2024gpt4ocard, geminiteam2024geminifamilyhighlycapable}. To further enhance the reasoning ability of MLLMs, increasing efforts have focused on improving MLLMs in multimodal reasoning. The proprietery model, o1 \cite{openai2024gpto1card} first realize strong multimodal reasoning with reasoning process similar to the Chain-of-Thought \cite{chain-of-thought} and other following works have also explored the multimodal reasoning either through training \cite{wu2024v,qi2024cogcom,shao2024visual} or prompting \cite{zhang2023makes,zhang2024prompt,zheng2023ddcot}.

\paragraph{Multimodal Reasoning Benchmarks} To evaluate the reasoning capabilities of MLLMs, numerous benchmarks have been developed with various focuses. Broad-coverage benchmarks such as MM-Bench \cite{liu2024mmbench}, MMMU \cite{yue2023mmmu} and MM-Vet \cite{yu2024mmvetevaluatinglargemultimodal} cover comprehensive reasoning challenges in real life scenarios, offering holistic insights into model performance. Others are developed with focuses on specific perspectives, such as TextVQA \cite{singh2019towards}, POPE \cite{li2023evaluatingobjecthallucinationlarge} and MATHVERSE \cite{zhang2024mathversedoesmultimodalllm} respectively challenge models with tasks in domains of reasoning about text, objects, mathematics in multimodal contexts. Recently, additional benchmarks have emerged targeting artificially created multipanel images—such as posters and screenshots—that combine several subfigures in structured layouts \cite{fan2024muffin,hsiao2025screenqalargescalequestionanswerpairs}, which require models to analyze spatial relationships and hierarchical structures in complex visual contexts.
However, current multi-modal benchmarks assume visual-text alignment, overlooking detecting critical errors of vision-language inconsistency in the input - a key challenge in real-world scenarios. Instead, we evaluate MLLMs’ ability to detect and localize such inconsistency via the proposed MMIR benchmark.

\paragraph{Inconsistency Checking}
Existing works on tasks related to checking or verifying inconsistency in the input are primarily in the language domain. For example, fact-checking \cite{thorne2018fever} requires a model to first retrieve evidence and then decide if a claim is supported, where the model must reason if contradictive information existed in the retrieved corpus. One step further, summary inconsistency detection \cite{laban2022summac} focuses on flagging any errors in summaries that create contradictions regardless of correctness, including incorrect use or hallucination of entities. As modern language models prosper, inconsistencies are found existing within their outputs \cite{ravichander2020systematicity} and across different outputs of paraphrased queries \cite{elazar2021measuring}, and efforts have been made towards the evaluation of those inconsistencies \cite{fabbri2021qafacteval,wang2020asking,lattimer2023fast}. In our research, we lead efforts in detecting inconsistencies in the field of vision and language.

\section{MMIR}

\begin{figure*}[t]
\setlength\tabcolsep{0pt}
\setlength{\abovecaptionskip}{0.1cm}
    \centering
    \includegraphics[width=\textwidth]{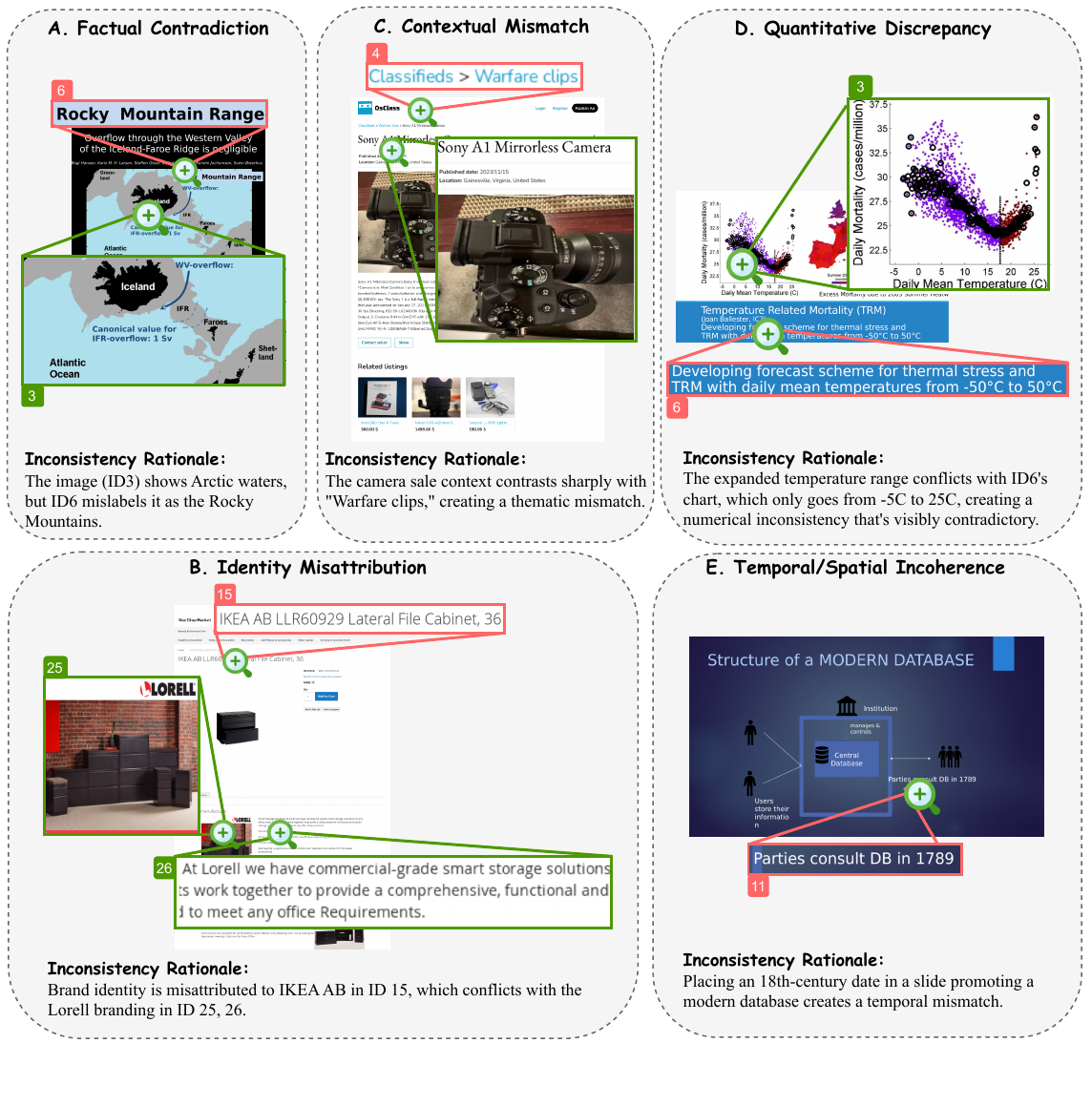}
    \vspace{-20pt}
    \caption{There are five inconsistency categories in the MMIR benchmark, posing diverse challenges.}
    \label{fig:MMIR_examples}
\end{figure*}

The MMIR benchmark is designed to assess how effectively MLLMs can detect and localize semantic mismatches within complex, layout-rich artifacts. Unlike conventional benchmarks that assume coherent visual–textual inputs, MMIR challenges models with realistic errors that require deep, cross-modal reasoning. In MMIR, errors are defined and categorized along five semantic dimensions:

\textit{A. Factual Contradiction}: Direct conflict between two elements (text–text, text–image, or image–image) within the artifact.

\textit{B. Identity Misattribution}: Mislabeling of entities (objects, locations, brands, people) that conflict with other elements.
    
\textit{C. Contextual Mismatch}: Tonal, thematic, or situational incompatibility between elements.

\textit{D. Quantitative Discrepancy}: Numerical or statistical inconsistencies between elements.

\textit{E. Temporal/Spatial Incoherence}: Implied timelines, dates, or spatial relationships that are impossible or conflicting.

Figure~\ref{fig:MMIR_examples} provides one example from each error type across web, office, and poster artifacts, illustrating the diverse challenges MMIR poses. 
Detailed definitions and examples of inconsistency error types can be found in Appendix~\ref{appendix:sec:error category}.

\subsection{Data Curation}

\begin{figure}[htbp]
\setlength\tabcolsep{0pt}
\setlength{\abovecaptionskip}{0.1cm}
    \centering
    \includegraphics[width=0.5\textwidth]{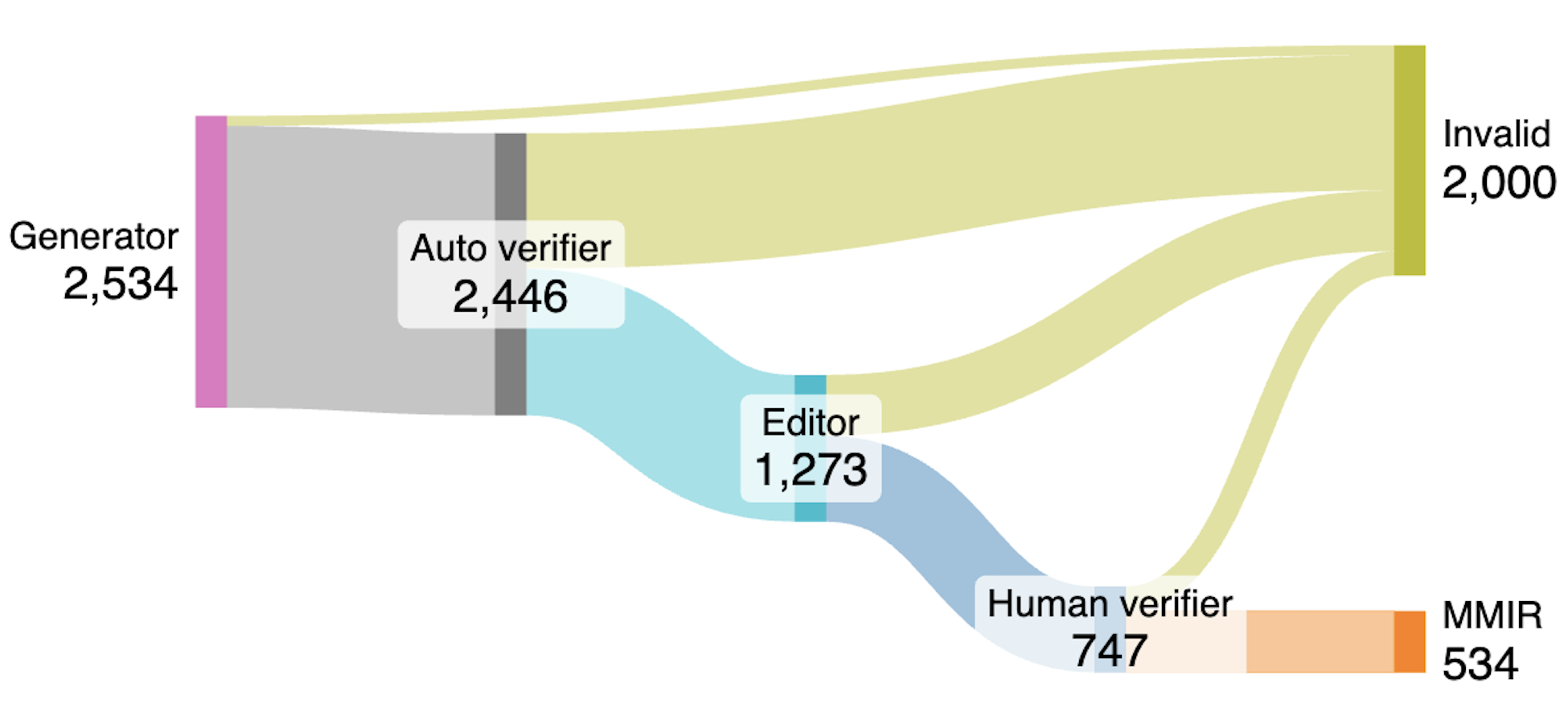}
    \caption{MMIR Data filtering process.}
    \label{fig:data_filter}
\end{figure}

MMIR’s data is curated through a four-stage pipeline (Figure~\ref{fig:data_filter}), ensuring high-quality, diverse, and challenging test cases.

\paragraph{Artifact Collection and Parsing}
We begin by manually selecting a total of 521 original artifacts from two domains: 349 webpages (sub-categories: shopping, classifieds, wiki) from VisualWebArena~\cite{Koh2024VisualWebArenaEM} and 172 presentations from Zenodo~\cite{zenodo}, categorized into Office (sub-categories: slides, charts, diagrams) and Posters. 
Each artifact $A_i$ is parsed using either using Document Object Model (DOM) or the \texttt{python-pptx} library to extract a set of elements $E_i = \{e_j\}_{j=1}^{n_i}$, where each element $e_j$is assigned a unique ID \(\mathtt{id}_j\) and labeled with its type, content, and a bounding box showing location information. Additionally, each artifact is paired with a Set-of-Marks (SoM) annotation $A_{i}^{\text{SoM}}$ derived from $E_i$. This structured metadata forms the basis for subsequent error injection and question-answer curation.

\paragraph{Synthetic Inconsistency Generation}
To simulate real-world errors, we prompt an MLLM, o1-1217~\cite{openai2024gpto1card}, as a generator with the annotated artifact and its element set $\{A_{i}^{\text{SoM}}, E_i\}$.
The generator produces 2,534 proposals,
each comprising a formatted edit instruction,
the ground-truth element or element pair introducing the inconsistency:
$$\mathtt{GT} \in \{\mathtt{id}_j\} \cup \{(\mathtt{id}_j, \mathtt{id}_k) | j \neq k\},$$
the inconsistency error type, and the accompanying rationale. Following a self-evaluation loop (details in Appendix~\ref{appendix:sec:generator}), 2,446 valid proposals are retained.

\paragraph{Automated Editing and Human Verification}
An auto-verification process then filters these proposals based on format and backend constraints (e.g., ensuring the target elements are editable), reducing the candidate set to 1,273, and saves low-level edit details, such as the path of the new image for an image edit, as inputs to the editor. 

An automated editor-implemented using the Chrome DevTools Protocol (CDP) for web pages and \texttt{python-pptx} for presentations-executes the approved edits, generating for each successful operation a modified pair: $\{A'_i, E'_i\}$ where $A'_i$ represents the modified artifact and $E'_i$ contains the updated element metadata after the edit.
For each pair, a descriptive caption set $C_i$ is generated, where each caption within $C_j$ details the element ID, location, and content summary of $e'_j$. These captions serve as references for later evaluation.
More details on the verifier and editor are provided in Appendix~\ref{appendix:sec:verifier and editor}.

\begin{table}[t]
\caption{\textbf{MMIR Statistics.} Breakdown of the dataset by artifact category and error type.}
\centering
\resizebox{\linewidth}{!}{
\begin{tabular}{lcr}
\toprule
\textbf{Category} & \textbf{\#Questions} & \textbf{Ave. \#Elements} \\
\midrule
\multicolumn{3}{l}{\textbf{Artifact Categories}} \\
\hspace{3mm} Web                & 240   & 38.8 \\
\hspace{6mm} - Shopping         & 108   & 46.1 \\
\hspace{6mm} - Wiki             & 28    & 44.9 \\
\hspace{6mm} - Classifieds      & 104   & 29.5 \\
\hspace{3mm} Office             & 223   & 9.1 \\
\hspace{6mm} - Slides           & 102   & 9.4 \\
\hspace{6mm} - Tables/Charts    & 61    & 4.1 \\
\hspace{6mm} - Diagrams         & 60    & 13.9 \\
\hspace{3mm} Poster             & 71    & 27.6 \\
\textbf{Total}                  & 543   & 24.9 \\

\midrule
\multicolumn{3}{l}{\textbf{Error Categories}} \\
\hspace{3mm} Factual Contradiction          & 138   & -- \\
\hspace{3mm} Identity Misattribution        & 84    & -- \\
\hspace{3mm} Contextual Mismatch            & 141   & -- \\
\hspace{3mm} Quantitative Discrepancy       & 76    & -- \\
\hspace{3mm} Temporal/Spatial Incoherence   & 95    & -- \\
\textbf{Total}                              & 543   & -- \\
\bottomrule
\end{tabular}
}
\label{tab:statistics}
\end{table}

Finally, human experts review 747 edited samples, resulting in a final dataset of 534 validated quintuples: 
$D_{MMIR} = \{A'_i, E'_i, \mathtt{GT}_i, \mathtt{category}_i, \mathtt{rationale}_i\}_{i=1}^{534}$, ensuring that only realistic and challenging samples remain. Table~\ref{tab:statistics} provides a detailed breakdown by artifact type, subcategory, and error type. For example, webpages are further divided into shopping, wiki, and classifieds, each with its average number of elements, while errors are distributed across the five defined categories. Notably, the average word count in multiple-choice questions is 382.6, whereas open-ended responses are fixed at 59 words.

\subsection{Evaluation}

MMIR assesses a model's ability to \emph{detect inconsistency}, i.e., identifying and localizing semantic mismatches where elements deviate from their expected roles within an artifact. To assess the model's performance comprehensively, each of the 534 test samples is provided to models under two distinct settings:

\paragraph{Open-Ended Setting}
Models receive the artifact $A'_i$ with a fixed prompt $Q_\text{open\_ended}$ and generate a free-form response that identifies the semantic mismatch. 
This formulation evaluates the model's ability to detect inconsistencies without relying on predefined answer options, thereby testing its unsupervised perception and reasoning.

\paragraph{Multiple-Choice Setting}
Models receive the artifact $A'_i$, but now with a combined prompt 
$
Q_{\text{MCQ}} = (Q_\text{open\_ended}, C_i).
$
Each candidate in $C_i$ is a textual description of an element. The model must select, from these options, the element(s) corresponding to the introduced inconsistency.

\paragraph{Evaluation Setup} 
For the MCQ setting, we utilize regular expressions to compare the MLLM's predicted answers against the ground truth, using accuracy as our metric. 
For the open-ended setting, o1-mini (0912) is employed as an LLM judge~\cite{hsu2023gpt,hackl2023gpt,liu2023g} to map the model's free-form response back to the most likely ground-truth element IDs.
The predicted IDs are then compared against $\mathtt{GT}_i$ to calculate accuracy.

\section{Experiments and Analysis}

\begin{table*}[t]
\caption{\textbf{The accuracy of six MLLMs under the two evaluation settings.} Proprietary models demonstrate higher performance as well as larger performance gain in the MCQ setting.}
\begin{center}
\vspace{-10pt}
\resizebox{0.75\linewidth}{!}{
\begin{tabular}{lcccc|cccc}
\toprule

& \multicolumn{4}{c}{\textbf{Open-ended}} & \multicolumn{4}{c}{\textbf{Multiple-choice}} \\
\cmidrule(l){2-5}
\cmidrule(l){6-9}

Models & Web & Office & Poster & Overall & Web & Office & Poster & Overall\\

\midrule
\multicolumn{9}{>{\columncolor{gray!15}}l}{\textit{Proprietary Models}}\\
o1 (1217)         & 47.91 & 59.19 & 38.73 & 51.40 & 47.91 & 58.52 & 46.47 & 52.15 \\
GPT-4o (1120)     & 25.00 & 42.60 & 30.98 & 33.14 & 37.29 & 58.96 & 47.88 & 47.75 \\
\multicolumn{9}{>{\columncolor{gray!15}}l}{\textit{Open-sourced Models}}\\
Qwen2.5-VL-72B    & 18.33 & 40.80 & 14.78 & 27.24 & 33.33 & 44.39 & 34.50 & 38.10 \\ 
Llama-3.2-90B-Vision-Instruct    & 7.08 & 23.76 & 7.04 & 14.04 & 20.62 & 23.31 & 29.57 & 22.94 \\ 

Qwen2.5-VL-7B     & 8.54  & 29.14 & 11.97 & 17.60 & 14.37 & 33.18 & 16.90 & 22.56 \\  
LLaVA-NeXT-7B     & 10.20 & 21.97 & 7.04  & 14.70 & 11.45 & 25.33 & 5.63  & 16.47 \\
InternVL2.5-8B    & 7.70  & 24.21 & 4.92  & 14.23 & 9.37  & 23.54 & 11.97 & 15.63 \\
Phi-3.5-Vision-4B & 6.87  & 24.43 & 7.04  & 14.23 & 1.66  & 8.52  & 0.00  & 4.30  \\
\bottomrule
\end{tabular}
}
\end{center}
\label{tab:main_eval_result}
\end{table*}

We first evaluate the advanced multimodal reasoning model o1~\cite{openai2024gpto1card} and seven other state-of-the-art MLLMs: GPT-4o~\cite{openai2024gpt4ocard}, Qwen2.5-VL~\cite{Qwen2.5-VL}, Llama-3.2~\cite{grattafiori2024llama}, LLaVA-NeXT~\cite{liu2024llavanext}, InternVL2.5~\cite{Chen2024ExpandingPB} and Phi-3.5-Vision~\cite{abdin2024phi3technicalreporthighly} on the MMIR benchmark. 
We implement open-source models using their default settings and select the 1217 version of o1 and the 1120 version of GPT-4o for evaluation. Model implementation details are provided in Appendix~\ref{appendix:sec:model_detail}.
We then examine error patterns across different inconsistency types and layout complexities, and finally explore how prompting strategies affect multimodal reasoning under the open-ended setting.

\subsection{Main Results}

As shown in Table~\ref{tab:main_eval_result}, proprietary models (o1 and GPT-4o) significantly outperform open-source alternatives, though all models exhibit substantial room for improvement. Appendix~\ref{appendix:sec:example_main settings} shows a qualitative example with question-answer and model response with qualitative analysis results.

\paragraph{Performance Gap Between Reasoning, Proprietary and Open-Source Models.} 
In both open-ended and MCQ settings, the reasoning o1 model substantially outperforms the rest, surpassing all 7B open-source models by over 24\%. 
Qwen2.5-VL-72B performs best among the tested open-source models, with an overall accuracy of 27.24\% in the vanilla setting.
The other proprietary model, GPT-4o, although missing the explicit reasoning ability of o1, outperforms open-source alternatives, reflecting stronger multimodal alignment and reasoning capabilities.

\paragraph{Impact of Semantic Cues.} 
GPT-4o sees a 14.61\% accuracy boost in the MCQ setting with additional element descriptions as options, narrowing its gap with o1 from 18.26\% to just 4.4\%. This indicates that GPT-4o relies heavily on semantic context when available. Similar accuracy boosts with additional semantic cues can be witnessed on Qwen2.5-VL-72B and Llama-3.2-90B-Vision-Instruct.

\paragraph{Inconsistent Gains for Open-Source Models.} 
Open-source models gain moderate or little accuracy when provided with MCQ-style prompts compared to proprietary  GPT-4o. 
Phi-3.5-Vision-4B experiences a 9.93\% drop, suggesting weaker reasoning capacity and less effective use of textual cues. 
The gap between proprietary and open-source models widens further in MCQ, highlighting the persistent challenge of integrating perceptual grounding with logical inference. 

\subsection{Error Analysis}

\begin{figure*}[h]
\setlength\tabcolsep{0pt}
\setlength{\abovecaptionskip}{0.1cm}
    \centering
    \includegraphics[width=0.9\textwidth]{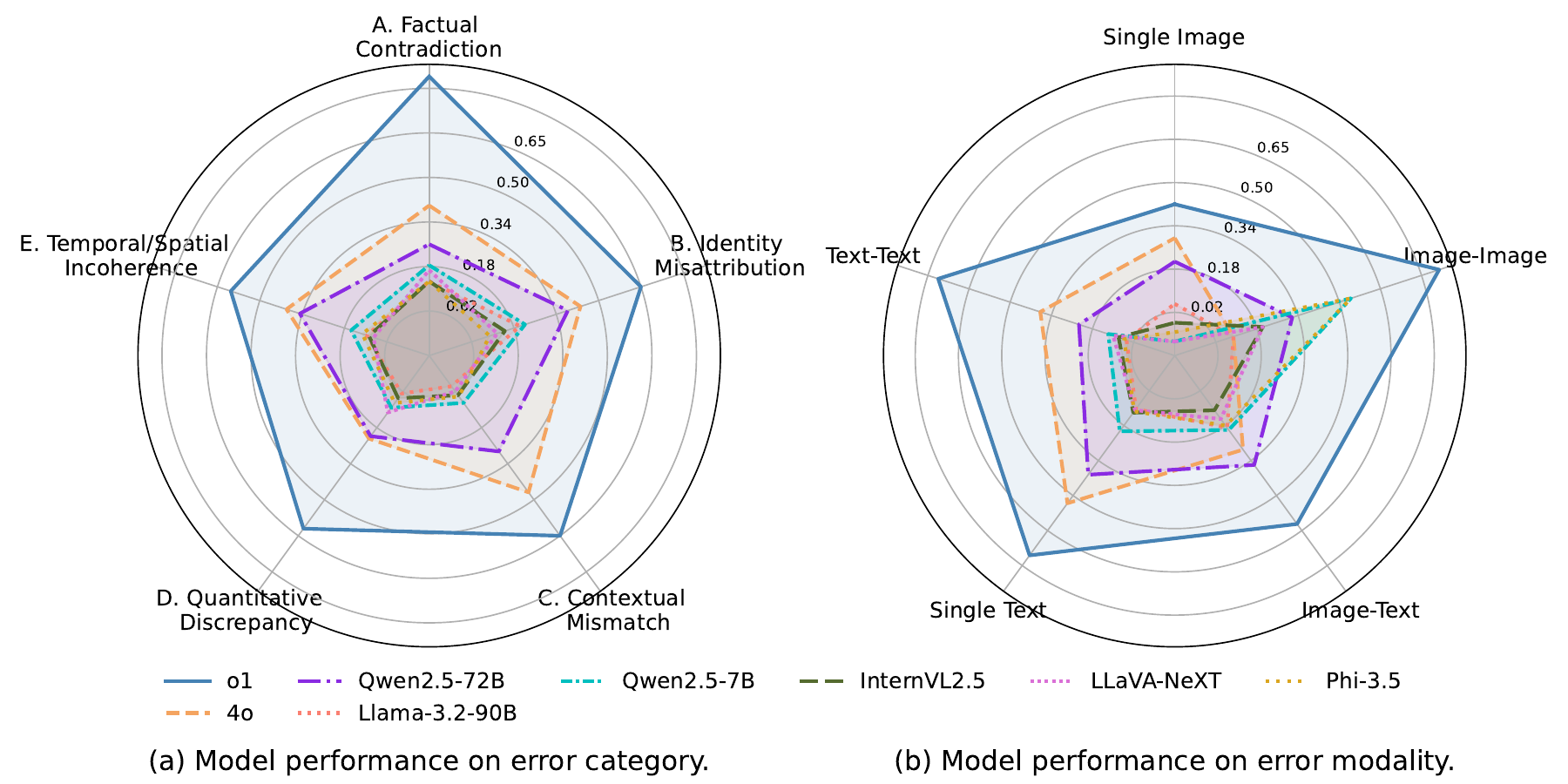}
    \caption{Fine-grained analysis of model performance.}
    \label{fig:ablation_error}
\end{figure*}

\subsubsection{Results Across Inconsistency Categories and Modalities}

To investigate how different types of inconsistencies affect model performance, we show the results across the category and modality of inconsistency in Figure~\ref{fig:ablation_error}. 

\paragraph{Inconsistency Categories}
Figure~\ref{fig:ablation_error}(a) breaks down accuracy by the five inconsistency error categories. Proprietary models (o1, GPT-4o) outperform open-source models across the board, but the gap is particularly pronounced for \emph{Factual Contradictions}, implying that high-capacity models may have stronger factual grounding and entity recognition. 
Interestingly, \emph{Quantitative Discrepancy} poses a substantial challenge for all models, highlighting a limitation in reasoning about numerical misalignment.

\paragraph{Inconsistency Modalities}
In Figure~\ref{fig:ablation_error}(b), we examine how accuracy varies by the modality of the inconsistency. Overall, inter-modality errors yield the highest performance, with image-image inconsistencies proving especially tractable. 
Next in difficulty are errors involving textual elements, including intra-modality errors (image-text) and errors posed by inconsistent textual elements, which require partial cross-modal and cross-element integration but can still leverage textual anchors. 
Finally, single image modality (those involving only one image field) inconsistencies pose the greatest challenge, as they demand more advanced visual understanding and the ability to reconcile contextual inconsistency without contradiction with another element.
These findings highlight that while models cope relatively well with pairwise conflicts, their capacity for deep visual or contextual reasoning remains underdeveloped.

\subsubsection{Impact of Layout Complexity}

\begin{figure}[t]
\setlength\tabcolsep{0pt}
\setlength{\abovecaptionskip}{0.1cm}
    \centering
    \includegraphics[width=\columnwidth]{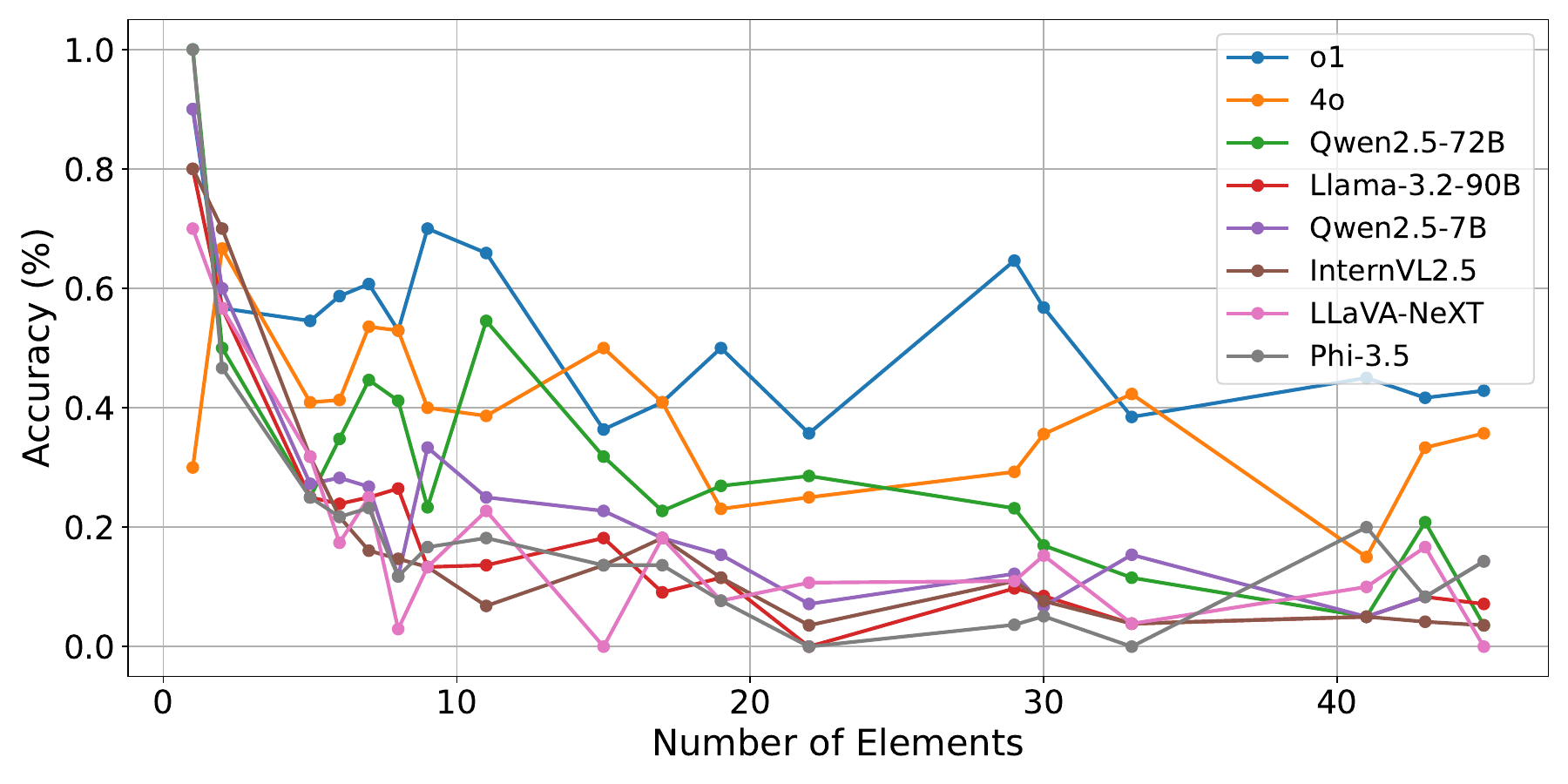}
    \caption{Model performance on layout complexity.}
    \label{fig:ablation_count}
\end{figure}

We further examine the relationship between model accuracy and the number of elements in an artifact. To ensure statistical significance, we only include data points where at least 10 samples share the same element count. As shown in Figure~\ref{fig:ablation_count}, the overall trend suggests that handling visually dense, information-rich artifacts remains a major challenge for current MLLMs.
(1) Performance declines sharply as the number of elements increases, highlighting the difficulty in parsing cluttered layouts.
(2) Proprietary models maintain higher accuracy in simpler layouts but degrade similarly in highly dense artifacts, indicating limitations in spatial reasoning. Open-source models struggle even in low-complexity settings, reinforcing the gap in perception and layout-aware inference.

\subsection{Probing on Prompting Methods}

\begin{table}[t]
\caption{\textbf{Probing results of different prompting methods.} Performance of each prompting method is directly compared with the vanilla setting. Gains are in blue and drops are in red.
}
\begin{center}
\vspace{-10pt}
\resizebox{\linewidth}{!}{
\begin{tabular}{lccccc}
\toprule

Models & Vanilla & + CoT & + SoM & + Both & MM-CoT\\
\midrule
\multicolumn{6}{>{\columncolor{gray!15}}l}{\textit{Proprietary Models}}\\
o1 (1217)           & 51.40  & -- & \textcolor{red}{-0.66}  & -- & \textcolor{blue}{+0.09}\\
GPT-4o (1120)       & 33.14  & -- & \textcolor{blue}{+5.34} & -- & \textcolor{blue}{+4.40}\\
\multicolumn{6}{>{\columncolor{gray!15}}l}{\textit{Open-sourced Models}}\\
Qwen2.5-VL-7B       & 17.60 & \textcolor{blue}{+0.28} & \textcolor{blue}{+0.09} & \textcolor{blue}{+0.28} & \textcolor{blue}{+4.59}\\ 
LLaVA-NeXT-7B       & 14.70 & \textcolor{red}{-1.78}  & \textcolor{red}{-2.53} & \textcolor{red}{-0.47} & \textcolor{blue}{+3.65}\\
InternVL2.5-8B      & 14.23 & \textcolor{blue}{+2.24} & \textcolor{red}{-0.66} & \textcolor{red}{-1.41} & \textcolor{red}{-0.85}\\
Phi-3.5-Vision-4B   & 14.23 & \textcolor{red}{-0.38}  & \textcolor{blue}{+0.47} & \textcolor{blue}{+0.84} & \textcolor{blue}{+0.65}\\
\bottomrule
\end{tabular}
}
\end{center}
\label{tab:ablation_cot_som}
\end{table}

We further investigate whether textual or visual prompts can alleviate the reasoning bottleneck. Table~\ref{tab:ablation_cot_som} compares \emph{Chain-of-Thought (CoT)} prompting~\cite{chain-of-thought} and \emph{Set-of-Mark (SoM)} visual augmentation~\cite{Yang2023SetofMarkPU}, as well as their combination on six of the tested models. We also explored an interleaved multimodal reasoning strategy, which we term \emph{Multimodal Interleaved CoT (MM-CoT)} to further integrate and refine reasoning across both visual and textual modalities.

\subsubsection{Chain-of-Thought (CoT) Prompting}
To assess whether explicit reasoning instructions can enhance performance, we apply CoT prompting~\cite{chain-of-thought} to the four open-sourced models (benchmarked proprietary models have API guides to not include additional CoT prompting). CoT prompting is a technique that encourages models to solve complex problems by generating intermediate reasoning steps, thereby enhancing their problem-solving abilities. Inspired by the common CoT setup, we append a text prompt: “Think step by step first, then provide the final answer.” to each base input in this ablation setting.

As shown in Table~\ref{tab:ablation_cot_som}, CoT prompting yields negligible or even negative effects on accuracy. This suggests that simply injecting explicit reasoning steps is insufficient when the underlying model lacks strong cross-modal alignment or robust logical inference mechanisms.

\subsubsection{Set-of-Mark (SoM) Prompting}

\begin{figure}[h]
\setlength\tabcolsep{0pt}
\setlength{\abovecaptionskip}{0.1cm}
    \centering
    \includegraphics[width=\columnwidth]{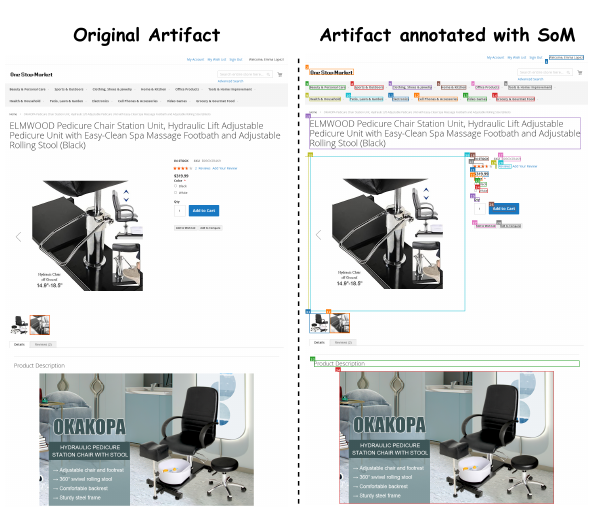}
    \caption{Example of original artifact in MMIR (left) and artifact annotated with Set-of-Mark in the probing analysis (right).}
    \label{fig:som}
\end{figure}

We next examine the effect of SoM visual prompting~\cite{Yang2023SetofMarkPU}. SoM prompting is a visual prompting method that enhances the visual grounding abilities of LMMs by overlaying images with spatial and speakable marks, such as alphanumerics or boxes, to partition the image into regions at different levels of granularity. Following the original work, we use visual annotations (e.g., bounding boxes with numerical IDs) to highlight elements in the artifact, as shown in Figure~\ref{fig:som}. This is a visual analog to CoT, aiming to guide the model’s attention during inference.

The result shows that these additional visual cues yield moderate improvements for GPT-4o (5.34\%) yet confuse the rest of the models, leading to little or even slightly degraded performance, likely because the additional visual cues interfere with the model’s initial perception.

When combined with CoT prompting, SoM provides little gains for some open-source models but remains largely inconsistent or even detrimental for others. This indicates that simply stacking CoT and SoM techniques does not guarantee improved performance, underscoring the need for more sophisticated strategies to unify visual cues with explicit reasoning steps.

\subsubsection{Multimodal Interleaved CoT (MM-CoT)}

Our previous analyses indicate that single-modality prompts (CoT or SoM) often yield minimal or even detrimental gains in the open-ended setting when models receive no textual hints about which elements might be inconsistent. We hypothesize that MMIR tasks demand \textit{iterative} reasoning that tightly integrates both visual and textual modalities. To address this, we propose \emph{Multimodal Interleaved CoT (MM-CoT)}, a two-stage approach explicitly designed to weave visual cues into a step-by-step reasoning process:

\paragraph{Stage 1: Initial Candidate Generation}  
The model receives the same input in Stage 1 as in the open-ended setting, generating its top five predictions (along with associated reasoning). Using o1-mini (0912) to interpret these responses, we map each prediction back to one or a pair of element IDs from the artifact’s metadata \(C_i\). We then highlight the bounding boxes of those elements on the artifact image, producing an SoM-annotated version to be used in the next stage.

\paragraph{Stage 2: Multimodal Refinement}  
The model is subsequently given the SoM-annotated artifact from Stage 1, alongside the textual reasoning it generated previously. This additional visual context helps the model refine its earlier predictions, integrating both the visual bounding-box annotations and the initial textual reasoning to arrive at a final answer.

\paragraph{Results}  
As shown in Table~\ref{tab:ablation_cot_som}, MM-CoT outperforms all other prompting methods. GPT-4o, for example, improves by 4.40\% over its vanilla baseline, while open-source models gain an average of around 2\% improvements. These findings underscore the importance of iterative cross-modal reasoning: once textual inferences guide which visual elements to focus on, SoM annotations become more informative, and the overall reasoning process becomes more accurate. Although the bounding boxes used for SoM are derived from ground-truth references, this probing experiment demonstrates that \emph{interleaved} multimodal interaction is a promising direction for closing the reasoning gap in challenging, inconsistency-heavy scenarios.

\section{Discussion and Conclusion}

In this work, we introduce the \textit{Multimodal Inconsistency Reasoning Benchmark (MMIR)} to evaluate how well MLLMs detect and localize semantic mismatches in complex real-world artifacts. MMIR challenges models across five error categories and two reasoning settings for a detailed assessment of multimodal reasoning.
Our experiments show that even advanced proprietary models struggle with open-ended inconsistency detection. Although providing natural-language descriptions in a multiple-choice format offers modest gains, standard prompting techniques (e.g., Chain-of-Thought and Set-of-Mark) yield inconsistent or negative effects, while a proposed Multimodal Interleaved CoT (MM-CoT) method that iteratively refines reasoning by integrating visual and textual modalities, yielding greater performance improvements. Despite these advances, significant challenges remain, motivating further research on robust multimodal reasoning for real-world inconsistency detection.

\section*{Limitations}

While MMIR provides a rigorous framework for evaluating multimodal inconsistency reasoning, it is not without its limitations. Annotating and verifying inconsistencies in layout-rich artifacts remains a labor-intensive process. Although MMIR’s pipeline integrates automated editing and verification, the overall scale is still limited by the need for careful human review.
Although these domains capture a range of layouts and content types, they do not encompass the full variety of real-world multimodal artifacts (e.g., multi-page documents, social media feeds, or mobile application interfaces). 
On the other hand, synthetic error generation—while effective for systematically introducing controlled inconsistencies—may not perfectly mirror the nuanced mistakes that occur in human-generated content. This could lead to discrepancies between model performance on MMIR and in truly open-ended, real-world scenarios.
Scaling up the dataset to cover broader domains, more intricate layouts, and diverse error types would strengthen its ability to serve as a comprehensive benchmark for real-world multimodal inconsistency detection.

\section*{Acknowledgments}
This research project is partially sponsored by an eBay Research Award and has benefited from the Microsoft Accelerate Foundation Models Research (AFMR) grant program.

\bibliography{main}

\begin{thebibliography}{52}
\providecommand{\natexlab}[1]{#1}

\bibitem[{Abdin et~al.(2024)Abdin, Aneja, Awadalla, Awadallah, Awan, Bach, Bahree, Bakhtiari, Bao, Behl, Benhaim, Bilenko, Bjorck, Bubeck, Cai, Cai, Chaudhary, Chen, Chen, Chen, Chen, Chen, Cheng, Chopra, Dai, Dixon, Eldan, Fragoso, Gao, Gao, Gao, Garg, Giorno, Goswami, Gunasekar, Haider, Hao, Hewett, Hu, Huynh, Iter, Jacobs, Javaheripi, Jin, Karampatziakis, Kauffmann, Khademi, Kim, Kim, Kurilenko, Lee, Lee, Li, Li, Liang, Liden, Lin, Lin, Liu, Liu, Liu, Liu, Liu, Luo, Madan, Mahmoudzadeh, Majercak, Mazzola, Mendes, Mitra, Modi, Nguyen, Norick, Patra, Perez-Becker, Portet, Pryzant, Qin, Radmilac, Ren, de~Rosa, Rosset, Roy, Ruwase, Saarikivi, Saied, Salim, Santacroce, Shah, Shang, Sharma, Shen, Shukla, Song, Tanaka, Tupini, Vaddamanu, Wang, Wang, Wang, Wang, Wang, Wang, Ward, Wen, Witte, Wu, Wu, Wyatt, Xiao, Xu, Xu, Xu, Xue, Yadav, Yang, Yang, Yang, Yang, Yu, Yuan, Zhang, Zhang, Zhang, Zhang, Zhang, Zhang, Zhang, and Zhou}]{abdin2024phi3technicalreporthighly}
Marah Abdin, Jyoti Aneja, Hany Awadalla, Ahmed Awadallah, Ammar~Ahmad Awan, Nguyen Bach, Amit Bahree, Arash Bakhtiari, Jianmin Bao, Harkirat Behl, Alon Benhaim, Misha Bilenko, Johan Bjorck, Sébastien Bubeck, Martin Cai, Qin Cai, Vishrav Chaudhary, Dong Chen, Dongdong Chen, Weizhu Chen, Yen-Chun Chen, Yi-Ling Chen, Hao Cheng, Parul Chopra, Xiyang Dai, Matthew Dixon, Ronen Eldan, Victor Fragoso, Jianfeng Gao, Mei Gao, Min Gao, Amit Garg, Allie~Del Giorno, Abhishek Goswami, Suriya Gunasekar, Emman Haider, Junheng Hao, Russell~J. Hewett, Wenxiang Hu, Jamie Huynh, Dan Iter, Sam~Ade Jacobs, Mojan Javaheripi, Xin Jin, Nikos Karampatziakis, Piero Kauffmann, Mahoud Khademi, Dongwoo Kim, Young~Jin Kim, Lev Kurilenko, James~R. Lee, Yin~Tat Lee, Yuanzhi Li, Yunsheng Li, Chen Liang, Lars Liden, Xihui Lin, Zeqi Lin, Ce~Liu, Liyuan Liu, Mengchen Liu, Weishung Liu, Xiaodong Liu, Chong Luo, Piyush Madan, Ali Mahmoudzadeh, David Majercak, Matt Mazzola, Caio César~Teodoro Mendes, Arindam Mitra, Hardik Modi, Anh Nguyen,
  Brandon Norick, Barun Patra, Daniel Perez-Becker, Thomas Portet, Reid Pryzant, Heyang Qin, Marko Radmilac, Liliang Ren, Gustavo de~Rosa, Corby Rosset, Sambudha Roy, Olatunji Ruwase, Olli Saarikivi, Amin Saied, Adil Salim, Michael Santacroce, Shital Shah, Ning Shang, Hiteshi Sharma, Yelong Shen, Swadheen Shukla, Xia Song, Masahiro Tanaka, Andrea Tupini, Praneetha Vaddamanu, Chunyu Wang, Guanhua Wang, Lijuan Wang, Shuohang Wang, Xin Wang, Yu~Wang, Rachel Ward, Wen Wen, Philipp Witte, Haiping Wu, Xiaoxia Wu, Michael Wyatt, Bin Xiao, Can Xu, Jiahang Xu, Weijian Xu, Jilong Xue, Sonali Yadav, Fan Yang, Jianwei Yang, Yifan Yang, Ziyi Yang, Donghan Yu, Lu~Yuan, Chenruidong Zhang, Cyril Zhang, Jianwen Zhang, Li~Lyna Zhang, Yi~Zhang, Yue Zhang, Yunan Zhang, and Xiren Zhou. 2024.
\newblock \href {https://arxiv.org/abs/2404.14219} {Phi-3 technical report: A highly capable language model locally on your phone}.
\newblock \emph{Preprint}, arXiv:2404.14219.

\bibitem[{Alayrac et~al.(2022)Alayrac, Donahue, Luc, Miech, Barr, Hasson, Lenc, Mensch, Millican, Reynolds et~al.}]{alayrac2022flamingo}
Jean-Baptiste Alayrac, Jeff Donahue, Pauline Luc, Antoine Miech, Iain Barr, Yana Hasson, Karel Lenc, Arthur Mensch, Katherine Millican, Malcolm Reynolds, et~al. 2022.
\newblock Flamingo: a visual language model for few-shot learning.
\newblock \emph{Advances in neural information processing systems}, 35:23716--23736.

\bibitem[{Bai et~al.(2023)Bai, Bai, Yang, Wang, Tan, Wang, Lin, Zhou, and Zhou}]{bai2023qwen}
Jinze Bai, Shuai Bai, Shusheng Yang, Shijie Wang, Sinan Tan, Peng Wang, Junyang Lin, Chang Zhou, and Jingren Zhou. 2023.
\newblock Qwen-vl: A versatile vision-language model for understanding, localization, text reading, and beyond.
\newblock \emph{arXiv preprint arXiv:2308.12966}, 1(2):3.

\bibitem[{Chen et~al.(2023)Chen, Zhang, Zeng, Zhang, Zhu, and Zhao}]{chen2023shikra}
Keqin Chen, Zhao Zhang, Weili Zeng, Richong Zhang, Feng Zhu, and Rui Zhao. 2023.
\newblock Shikra: Unleashing multimodal llm's referential dialogue magic.
\newblock \emph{arXiv preprint arXiv:2306.15195}.

\bibitem[{Chen et~al.(2024)Chen, Wang, Cao, Liu, Gao, Cui, Zhu, Ye, Tian, Liu, Gu, Wang, Li, Ren, Chen, Luo, Wang, Jiang, Wang, He, Shi, Zhang, Lv, Wang, Shao, Chu, Tu, He, Wu, Deng, Ge, Chen, Dou, Lu, Zhu, Lu, Lin, Qiao, Dai, and Wang}]{Chen2024ExpandingPB}
Zhe Chen, Weiyun Wang, Yue Cao, Yangzhou Liu, Zhangwei Gao, Erfei Cui, Jinguo Zhu, Shenglong Ye, Hao Tian, Zhaoyang Liu, Lixin Gu, Xuehui Wang, Qingyun Li, Yiming Ren, Zixuan Chen, Jiapeng Luo, Jiahao Wang, Tan Jiang, Bo~Wang, Conghui He, Botian Shi, Xingcheng Zhang, Han Lv, Yi~Wang, Wenqi Shao, Pei Chu, Zhongying Tu, Tong He, Zhiyong Wu, Hui Deng, Jiaye Ge, Kaiming Chen, Min Dou, Lewei Lu, Xizhou Zhu, Tong Lu, Dahu Lin, Yunfeng Qiao, Jifeng Dai, and Wenhai Wang. 2024.
\newblock \href {https://api.semanticscholar.org/CorpusID:274581884} {Expanding performance boundaries of open-source multimodal models with model, data, and test-time scaling}.
\newblock \emph{ArXiv}, abs/2412.05271.

\bibitem[{Chowdhery et~al.(2023)Chowdhery, Narang, Devlin, Bosma, Mishra, Roberts, Barham, Chung, Sutton, Gehrmann et~al.}]{chowdhery2023palm}
Aakanksha Chowdhery, Sharan Narang, Jacob Devlin, Maarten Bosma, Gaurav Mishra, Adam Roberts, Paul Barham, Hyung~Won Chung, Charles Sutton, Sebastian Gehrmann, et~al. 2023.
\newblock Palm: Scaling language modeling with pathways.
\newblock \emph{Journal of Machine Learning Research}, 24(240):1--113.

\bibitem[{Elazar et~al.(2021)Elazar, Kassner, Ravfogel, Ravichander, Hovy, Sch{\"u}tze, and Goldberg}]{elazar2021measuring}
Yanai Elazar, Nora Kassner, Shauli Ravfogel, Abhilasha Ravichander, Eduard Hovy, Hinrich Sch{\"u}tze, and Yoav Goldberg. 2021.
\newblock Measuring and improving consistency in pretrained language models.
\newblock \emph{Transactions of the Association for Computational Linguistics}, 9:1012--1031.

\bibitem[{{European Organization For Nuclear Research} and {OpenAIRE}(2013)}]{zenodo}
{European Organization For Nuclear Research} and {OpenAIRE}. 2013.
\newblock \href {https://doi.org/10.25495/7GXK-RD71} {Zenodo}.

\bibitem[{Fabbri et~al.(2021)Fabbri, Wu, Liu, and Xiong}]{fabbri2021qafacteval}
Alexander~R Fabbri, Chien-Sheng Wu, Wenhao Liu, and Caiming Xiong. 2021.
\newblock Qafacteval: Improved qa-based factual consistency evaluation for summarization.
\newblock \emph{arXiv preprint arXiv:2112.08542}.

\bibitem[{Fan et~al.(2024{\natexlab{a}})Fan, Ding, Kuo, Jiang, Zhao, Guan, Yang, Zhang, and Wang}]{fan2024read}
Yue Fan, Lei Ding, Ching-Chen Kuo, Shan Jiang, Yang Zhao, Xinze Guan, Jie Yang, Yi~Zhang, and Xin~Eric Wang. 2024{\natexlab{a}}.
\newblock Read anywhere pointed: Layout-aware gui screen reading with tree-of-lens grounding.
\newblock \emph{arXiv preprint arXiv:2406.19263}.

\bibitem[{Fan et~al.(2024{\natexlab{b}})Fan, Gu, Zhou, Yan, Jiang, Kuo, Zhao, Guan, and Wang}]{fan2024muffin}
Yue Fan, Jing Gu, Kaiwen Zhou, Qianqi Yan, Shan Jiang, Ching-Chen Kuo, Yang Zhao, Xinze Guan, and Xin Wang. 2024{\natexlab{b}}.
\newblock Muffin or chihuahua? challenging multimodal large language models with multipanel vqa.
\newblock In \emph{Proceedings of the 62nd Annual Meeting of the Association for Computational Linguistics (Volume 1: Long Papers)}, pages 6845--6863.

\bibitem[{Feng et~al.(2024)Feng, Zhu, Fu, Jampani, Akula, He, Basu, Wang, and Wang}]{feng2024layoutgpt}
Weixi Feng, Wanrong Zhu, Tsu-jui Fu, Varun Jampani, Arjun Akula, Xuehai He, Sugato Basu, Xin~Eric Wang, and William~Yang Wang. 2024.
\newblock Layoutgpt: Compositional visual planning and generation with large language models.
\newblock \emph{Advances in Neural Information Processing Systems}, 36.

\bibitem[{Floridi and Chiriatti(2020)}]{floridi2020gpt}
Luciano Floridi and Massimo Chiriatti. 2020.
\newblock Gpt-3: Its nature, scope, limits, and consequences.
\newblock \emph{Minds and Machines}, 30:681--694.

\bibitem[{Grattafiori et~al.(2024)Grattafiori, Dubey, Jauhri, Pandey, Kadian, Al-Dahle, Letman, Mathur, Schelten, Vaughan et~al.}]{grattafiori2024llama}
Aaron Grattafiori, Abhimanyu Dubey, Abhinav Jauhri, Abhinav Pandey, Abhishek Kadian, Ahmad Al-Dahle, Aiesha Letman, Akhil Mathur, Alan Schelten, Alex Vaughan, et~al. 2024.
\newblock The llama 3 herd of models.
\newblock \emph{arXiv preprint arXiv:2407.21783}.

\bibitem[{Guo et~al.(2025)Guo, Yang, Zhang, Song, Zhang, Xu, Zhu, Ma, Wang, Bi et~al.}]{guo2025deepseek}
Daya Guo, Dejian Yang, Haowei Zhang, Junxiao Song, Ruoyu Zhang, Runxin Xu, Qihao Zhu, Shirong Ma, Peiyi Wang, Xiao Bi, et~al. 2025.
\newblock Deepseek-r1: Incentivizing reasoning capability in llms via reinforcement learning.
\newblock \emph{arXiv preprint arXiv:2501.12948}.

\bibitem[{Hackl et~al.(2023)Hackl, M{\"u}ller, Granitzer, and Sailer}]{hackl2023gpt}
Veronika Hackl, Alexandra~Elena M{\"u}ller, Michael Granitzer, and Maximilian Sailer. 2023.
\newblock Is gpt-4 a reliable rater? evaluating consistency in gpt-4's text ratings.
\newblock In \emph{Frontiers in Education}, volume~8, page 1272229. Frontiers Media SA.

\bibitem[{Hsiao et~al.(2025)Hsiao, Zubach, Baechler, Sunkara, Carbune, Lin, Wang, Zhu, and Chen}]{hsiao2025screenqalargescalequestionanswerpairs}
Yu-Chung Hsiao, Fedir Zubach, Gilles Baechler, Srinivas Sunkara, Victor Carbune, Jason Lin, Maria Wang, Yun Zhu, and Jindong Chen. 2025.
\newblock \href {https://arxiv.org/abs/2209.08199} {Screenqa: Large-scale question-answer pairs over mobile app screenshots}.
\newblock \emph{Preprint}, arXiv:2209.08199.

\bibitem[{Hsu et~al.(2023)Hsu, Huang, Rossi, Kim, Giles, and Huang}]{hsu2023gpt}
Ting-Yao Hsu, Chieh-Yang Huang, Ryan Rossi, Sungchul Kim, C~Lee Giles, and Ting-Hao~K Huang. 2023.
\newblock Gpt-4 as an effective zero-shot evaluator for scientific figure captions.
\newblock \emph{arXiv preprint arXiv:2310.15405}.

\bibitem[{Koh et~al.(2024)Koh, Lo, Jang, Duvvur, Lim, Huang, Neubig, Zhou, Salakhutdinov, and Fried}]{Koh2024VisualWebArenaEM}
Jing~Yu Koh, Robert Lo, Lawrence Jang, Vikram Duvvur, Ming~Chong Lim, Po-Yu Huang, Graham Neubig, Shuyan Zhou, Ruslan Salakhutdinov, and Daniel Fried. 2024.
\newblock \href {https://api.semanticscholar.org/CorpusID:271915493} {Visualwebarena: Evaluating multimodal agents on realistic visual web tasks}.

\bibitem[{Kojima et~al.(2022)Kojima, Gu, Reid, Matsuo, and Iwasawa}]{kojima2022large}
Takeshi Kojima, Shixiang~Shane Gu, Machel Reid, Yutaka Matsuo, and Yusuke Iwasawa. 2022.
\newblock Large language models are zero-shot reasoners.
\newblock \emph{Advances in neural information processing systems}, 35:22199--22213.

\bibitem[{Laban et~al.(2022)Laban, Schnabel, Bennett, and Hearst}]{laban2022summac}
Philippe Laban, Tobias Schnabel, Paul~N Bennett, and Marti~A Hearst. 2022.
\newblock Summac: Re-visiting nli-based models for inconsistency detection in summarization.
\newblock \emph{Transactions of the Association for Computational Linguistics}, 10:163--177.

\bibitem[{Lattimer et~al.(2023)Lattimer, Chen, Zhang, and Yang}]{lattimer2023fast}
Barrett~Martin Lattimer, Patrick Chen, Xinyuan Zhang, and Yi~Yang. 2023.
\newblock Fast and accurate factual inconsistency detection over long documents.
\newblock \emph{arXiv preprint arXiv:2310.13189}.

\bibitem[{Li et~al.(2023{\natexlab{a}})Li, Li, Savarese, and Hoi}]{li2023blip}
Junnan Li, Dongxu Li, Silvio Savarese, and Steven Hoi. 2023{\natexlab{a}}.
\newblock Blip-2: Bootstrapping language-image pre-training with frozen image encoders and large language models.
\newblock In \emph{International conference on machine learning}, pages 19730--19742. PMLR.

\bibitem[{Li et~al.(2023{\natexlab{b}})Li, Du, Zhou, Wang, Zhao, and Wen}]{li2023evaluatingobjecthallucinationlarge}
Yifan Li, Yifan Du, Kun Zhou, Jinpeng Wang, Wayne~Xin Zhao, and Ji-Rong Wen. 2023{\natexlab{b}}.
\newblock \href {https://arxiv.org/abs/2305.10355} {Evaluating object hallucination in large vision-language models}.
\newblock \emph{Preprint}, arXiv:2305.10355.

\bibitem[{Liu et~al.(2024{\natexlab{a}})Liu, Li, Li, and Lee}]{liu2024improved}
Haotian Liu, Chunyuan Li, Yuheng Li, and Yong~Jae Lee. 2024{\natexlab{a}}.
\newblock Improved baselines with visual instruction tuning.
\newblock In \emph{Proceedings of the IEEE/CVF Conference on Computer Vision and Pattern Recognition}, pages 26296--26306.

\bibitem[{Liu et~al.(2024{\natexlab{b}})Liu, Li, Li, Li, Zhang, Shen, and Lee}]{liu2024llavanext}
Haotian Liu, Chunyuan Li, Yuheng Li, Bo~Li, Yuanhan Zhang, Sheng Shen, and Yong~Jae Lee. 2024{\natexlab{b}}.
\newblock \href {https://llava-vl.github.io/blog/2024-01-30-llava-next/} {Llava-next: Improved reasoning, ocr, and world knowledge}.

\bibitem[{Liu et~al.(2024{\natexlab{c}})Liu, Ou, Song, Qu, Lam, Xiong, Chen, Neubig, and Yue}]{liu2024harnessingwebpageuistextrich}
Junpeng Liu, Tianyue Ou, Yifan Song, Yuxiao Qu, Wai Lam, Chenyan Xiong, Wenhu Chen, Graham Neubig, and Xiang Yue. 2024{\natexlab{c}}.
\newblock \href {https://arxiv.org/abs/2410.13824} {Harnessing webpage uis for text-rich visual understanding}.
\newblock \emph{Preprint}, arXiv:2410.13824.

\bibitem[{Liu et~al.(2023)Liu, Iter, Xu, Wang, Xu, and Zhu}]{liu2023g}
Yang Liu, Dan Iter, Yichong Xu, Shuohang Wang, Ruochen Xu, and Chenguang Zhu. 2023.
\newblock G-eval: Nlg evaluation using gpt-4 with better human alignment.
\newblock \emph{arXiv preprint arXiv:2303.16634}.

\bibitem[{Liu et~al.(2024{\natexlab{d}})Liu, Duan, Zhang, Li, Zhang, Zhao, Yuan, Wang, He, Liu et~al.}]{liu2024mmbench}
Yuan Liu, Haodong Duan, Yuanhan Zhang, Bo~Li, Songyang Zhang, Wangbo Zhao, Yike Yuan, Jiaqi Wang, Conghui He, Ziwei Liu, et~al. 2024{\natexlab{d}}.
\newblock Mmbench: Is your multi-modal model an all-around player?
\newblock In \emph{European conference on computer vision}, pages 216--233. Springer.

\bibitem[{OpenAI(2024{\natexlab{a}})}]{openai2024gpt4ocard}
OpenAI. 2024{\natexlab{a}}.
\newblock \href {https://arxiv.org/abs/2410.21276} {Gpt-4o system card}.
\newblock \emph{Preprint}, arXiv:2410.21276.

\bibitem[{OpenAI(2024{\natexlab{b}})}]{openai2024gpto1card}
OpenAI. 2024{\natexlab{b}}.
\newblock \href {https://arxiv.org/abs/2412.16720} {Openai o1 system card}.
\newblock \emph{Preprint}, arXiv:2412.16720.

\bibitem[{Peng et~al.(2023)Peng, Wang, Dong, Hao, Huang, Ma, and Wei}]{peng2023kosmos}
Zhiliang Peng, Wenhui Wang, Li~Dong, Yaru Hao, Shaohan Huang, Shuming Ma, and Furu Wei. 2023.
\newblock Kosmos-2: Grounding multimodal large language models to the world.
\newblock \emph{arXiv preprint arXiv:2306.14824}.

\bibitem[{Qi et~al.(2024)Qi, Ding, Wang, Bai, Lv, Hong, Xu, Hou, Li, Dong et~al.}]{qi2024cogcom}
Ji~Qi, Ming Ding, Weihan Wang, Yushi Bai, Qingsong Lv, Wenyi Hong, Bin Xu, Lei Hou, Juanzi Li, Yuxiao Dong, et~al. 2024.
\newblock Cogcom: Train large vision-language models diving into details through chain of manipulations.
\newblock \emph{arXiv preprint arXiv:2402.04236}.

\bibitem[{Radford et~al.(2021)Radford, Kim, Hallacy, Ramesh, Goh, Agarwal, Sastry, Askell, Mishkin, Clark et~al.}]{radford2021learning}
Alec Radford, Jong~Wook Kim, Chris Hallacy, Aditya Ramesh, Gabriel Goh, Sandhini Agarwal, Girish Sastry, Amanda Askell, Pamela Mishkin, Jack Clark, et~al. 2021.
\newblock Learning transferable visual models from natural language supervision.
\newblock In \emph{International conference on machine learning}, pages 8748--8763. PMLR.

\bibitem[{Ravichander et~al.(2020)Ravichander, Hovy, Suleman, Trischler, and Cheung}]{ravichander2020systematicity}
Abhilasha Ravichander, Eduard Hovy, Kaheer Suleman, Adam Trischler, and Jackie Chi~Kit Cheung. 2020.
\newblock On the systematicity of probing contextualized word representations: The case of hypernymy in bert.
\newblock In \emph{Proceedings of the Ninth Joint Conference on Lexical and Computational Semantics}, pages 88--102.

\bibitem[{Shao et~al.(2024)Shao, Qian, Xiao, Song, Zong, Wang, Liu, and Li}]{shao2024visual}
Hao Shao, Shengju Qian, Han Xiao, Guanglu Song, Zhuofan Zong, Letian Wang, Yu~Liu, and Hongsheng Li. 2024.
\newblock Visual cot: Advancing multi-modal language models with a comprehensive dataset and benchmark for chain-of-thought reasoning.
\newblock In \emph{The Thirty-eight Conference on Neural Information Processing Systems Datasets and Benchmarks Track}.

\bibitem[{Singh et~al.(2019)Singh, Natarjan, Shah, Jiang, Chen, Parikh, and Rohrbach}]{singh2019towards}
Amanpreet Singh, Vivek Natarjan, Meet Shah, Yu~Jiang, Xinlei Chen, Devi Parikh, and Marcus Rohrbach. 2019.
\newblock Towards vqa models that can read.
\newblock In \emph{Proceedings of the IEEE Conference on Computer Vision and Pattern Recognition}, pages 8317--8326.

\bibitem[{Taori et~al.(2023)Taori, Gulrajani, Zhang, Dubois, Li, Guestrin, Liang, and Hashimoto}]{taori2023stanford}
Rohan Taori, Ishaan Gulrajani, Tianyi Zhang, Yann Dubois, Xuechen Li, Carlos Guestrin, Percy Liang, and Tatsunori~B Hashimoto. 2023.
\newblock Stanford alpaca: An instruction-following llama model.

\bibitem[{Team(2024)}]{geminiteam2024geminifamilyhighlycapable}
Gemini Team. 2024.
\newblock \href {https://arxiv.org/abs/2312.11805} {Gemini: A family of highly capable multimodal models}.
\newblock \emph{Preprint}, arXiv:2312.11805.

\bibitem[{Team(2025)}]{Qwen2.5-VL}
Qwen Team. 2025.
\newblock \href {https://qwenlm.github.io/blog/qwen2.5-vl/} {Qwen2.5-vl}.

\bibitem[{Thorne et~al.(2018)Thorne, Vlachos, Christodoulopoulos, and Mittal}]{thorne2018fever}
James Thorne, Andreas Vlachos, Christos Christodoulopoulos, and Arpit Mittal. 2018.
\newblock Fever: a large-scale dataset for fact extraction and verification.
\newblock \emph{arXiv preprint arXiv:1803.05355}.

\bibitem[{Touvron et~al.(2023)Touvron, Lavril, Izacard, Martinet, Lachaux, Lacroix, Rozi{\`e}re, Goyal, Hambro, Azhar et~al.}]{touvron2023llama}
Hugo Touvron, Thibaut Lavril, Gautier Izacard, Xavier Martinet, Marie-Anne Lachaux, Timoth{\'e}e Lacroix, Baptiste Rozi{\`e}re, Naman Goyal, Eric Hambro, Faisal Azhar, et~al. 2023.
\newblock Llama: Open and efficient foundation language models.
\newblock \emph{arXiv preprint arXiv:2302.13971}.

\bibitem[{Wang et~al.(2020)Wang, Cho, and Lewis}]{wang2020asking}
Alex Wang, Kyunghyun Cho, and Mike Lewis. 2020.
\newblock Asking and answering questions to evaluate the factual consistency of summaries.
\newblock \emph{arXiv preprint arXiv:2004.04228}.

\bibitem[{Wei et~al.(2022)Wei, Wang, Schuurmans, Bosma, Ichter, Xia, Chi, Le, and Zhou}]{chain-of-thought}
Jason Wei, Xuezhi Wang, Dale Schuurmans, Maarten Bosma, Brian Ichter, Fei Xia, Ed~H. Chi, Quoc~V. Le, and Denny Zhou. 2022.
\newblock Chain-of-thought prompting elicits reasoning in large language models.
\newblock In \emph{Proceedings of the 36th International Conference on Neural Information Processing Systems}, NIPS '22, Red Hook, NY, USA. Curran Associates Inc.

\bibitem[{Wu and Xie(2024)}]{wu2024v}
Penghao Wu and Saining Xie. 2024.
\newblock V?: Guided visual search as a core mechanism in multimodal llms.
\newblock In \emph{Proceedings of the IEEE/CVF Conference on Computer Vision and Pattern Recognition}, pages 13084--13094.

\bibitem[{Yang et~al.(2023)Yang, Zhang, Li, Zou, yue Li, and Gao}]{Yang2023SetofMarkPU}
Jianwei Yang, Hao Zhang, Feng Li, Xueyan Zou, Chun yue Li, and Jianfeng Gao. 2023.
\newblock \href {https://api.semanticscholar.org/CorpusID:266149987} {Set-of-mark prompting unleashes extraordinary visual grounding in gpt-4v}.
\newblock \emph{ArXiv}, abs/2310.11441.

\bibitem[{Yu et~al.(2024)Yu, Yang, Li, Wang, Lin, Liu, Wang, and Wang}]{yu2024mmvetevaluatinglargemultimodal}
Weihao Yu, Zhengyuan Yang, Linjie Li, Jianfeng Wang, Kevin Lin, Zicheng Liu, Xinchao Wang, and Lijuan Wang. 2024.
\newblock \href {https://arxiv.org/abs/2308.02490} {Mm-vet: Evaluating large multimodal models for integrated capabilities}.
\newblock \emph{Preprint}, arXiv:2308.02490.

\bibitem[{Yue et~al.(2024)Yue, Ni, Zhang, Zheng, Liu, Zhang, Stevens, Jiang, Ren, Sun, Wei, Yu, Yuan, Sun, Yin, Zheng, Yang, Liu, Huang, Sun, Su, and Chen}]{yue2023mmmu}
Xiang Yue, Yuansheng Ni, Kai Zhang, Tianyu Zheng, Ruoqi Liu, Ge~Zhang, Samuel Stevens, Dongfu Jiang, Weiming Ren, Yuxuan Sun, Cong Wei, Botao Yu, Ruibin Yuan, Renliang Sun, Ming Yin, Boyuan Zheng, Zhenzhu Yang, Yibo Liu, Wenhao Huang, Huan Sun, Yu~Su, and Wenhu Chen. 2024.
\newblock Mmmu: A massive multi-discipline multimodal understanding and reasoning benchmark for expert agi.
\newblock In \emph{Proceedings of CVPR}.

\bibitem[{Zhang et~al.(2024{\natexlab{a}})Zhang, Jiang, Zhang, Lin, Guo, Qiu, Zhou, Lu, Chang, Gao, and Li}]{zhang2024mathversedoesmultimodalllm}
Renrui Zhang, Dongzhi Jiang, Yichi Zhang, Haokun Lin, Ziyu Guo, Pengshuo Qiu, Aojun Zhou, Pan Lu, Kai-Wei Chang, Peng Gao, and Hongsheng Li. 2024{\natexlab{a}}.
\newblock \href {https://arxiv.org/abs/2403.14624} {Mathverse: Does your multi-modal llm truly see the diagrams in visual math problems?}
\newblock \emph{Preprint}, arXiv:2403.14624.

\bibitem[{Zhang et~al.(2023)Zhang, Zhou, and Liu}]{zhang2023makes}
Yuanhan Zhang, Kaiyang Zhou, and Ziwei Liu. 2023.
\newblock What makes good examples for visual in-context learning?
\newblock \emph{Advances in Neural Information Processing Systems}, 36:17773--17794.

\bibitem[{Zhang et~al.(2024{\natexlab{b}})Zhang, Qian, Peng, Liu, and Jia}]{zhang2024prompt}
Yuechen Zhang, Shengju Qian, Bohao Peng, Shu Liu, and Jiaya Jia. 2024{\natexlab{b}}.
\newblock Prompt highlighter: Interactive control for multi-modal llms.
\newblock In \emph{Proceedings of the IEEE/CVF Conference on Computer Vision and Pattern Recognition}, pages 13215--13224.

\bibitem[{Zheng et~al.(2023)Zheng, Yang, Tang, Zhou, and Yang}]{zheng2023ddcot}
Ge~Zheng, Bin Yang, Jiajin Tang, Hong-Yu Zhou, and Sibei Yang. 2023.
\newblock Ddcot: Duty-distinct chain-of-thought prompting for multimodal reasoning in language models.
\newblock \emph{Advances in Neural Information Processing Systems}, 36:5168--5191.

\end{thebibliography}

\appendix
\newpage

\section{Benchmark Details}
This appendix provides a comprehensive overview of the MMIR benchmark. It details the dataset curation process, including error category definitions, the synthetic inconsistency generation mechanism, the auto‐verification and human validation processes, and the task prompts for evaluation. These details are intended to facilitate reproducibility and provide clarity on the inner workings of MMIR.

\subsection{Inconsistency Error Category Definitions}
\label{appendix:sec:error category}

The MMIR benchmark employs five pre-defined error categories. These categories are designed based on semantic guidelines so that the generator model can propose diverse and generalizable inconsistencies without being tied to any specific artifact type.

\begin{itemize}
    \item \textbf{A. Factual Contradiction} 
    
    Direct conflict between two or more elements (text–text, text–image, or image–image) within the artifact.
    
    \emph{Example (Text–Text): The product title says “Caffeinated,” while the description states “Caffeine-free.”\\Example (Text–Image): The image shows a green tea bag, but the accompanying text describes a “fruit infusion.”}

    \item \textbf{B. Identity Misattribution} 
    
    Mislabeling of entities (objects, locations, brands, people) that conflict with other elements.
    
    \emph{Example: A product lists “Country of Origin: China” while the manufacturer is described as “Elmwood Inn (USA).”}

    \item \textbf{C. Contextual Mismatch} 
    
    Tonal, thematic, or situational incompatibility between elements.
    
    \emph{Example: A celebratory image of diplomats shaking hands is paired with an article about violent clashes.}

    \item \textbf{D. Quantitative Discrepancy} 
    
    Numerical or statistical inconsistencies between elements.
    
    \emph{Example: A graph labeled “50\% growth” shows flat bars.}

    \item \textbf{E. Temporal/Spatial Incoherence} 
    
    Implied timelines, dates, or spatial relationships that are impossible or conflicting.
    
    \emph{Example: A map labeled “North America” depicts landmarks from Europe.}
\end{itemize}

These definitions serve as guidelines during the synthetic inconsistency generation process, ensuring that the proposed errors are semantically meaningful and cover a broad spectrum of potential real-world mistakes.

\subsection{Generator Model and Self-Evaluation Loop}
\label{appendix:sec:generator}

\subsubsection{Generator Model Prompt}

To create adversarial examples, the generator model (o1, 1217) is provided with rich context consisting of the annotated artifact $A_i^{\text{SOM}}$ and its set of elements $E_i$. The task prompt includes detailed instructions regarding the types of modifications to propose, along with the following guidelines:

\begin{itemize}
    \item \textbf{Modification Format:} Each modification must be expressed as: 

    ```Modify [id] [original\_content] [new\_content]```

    For image fields, the original content includes the full details (e.g., URL), and the new content is a caption starting with "Image, description: ". For text fields, the new content should be of similar length to the original.

    \item \textbf{Error Categories:} The generator must propose one modification per error category. If it cannot propose an inconsistency for a given category, it may skip that category.
\end{itemize}

The generator output is structured as:

$$P_m=\left\{\text { edit }_m, \mathrm{GT}_m, \text { category }{ }_m, \text { rationale }_m\right\}$$

where the ground-truth $\mathrm{GT}_m$ is defined as:

$$\mathrm{GT}_m \in\left\{\mathrm{id}_j\right\} \cup\left\{\left(\mathrm{id}_j, \mathrm{id}_k\right) \mid j \neq k\right\}$$

indicating either a single-element ID (for single-element inconsistencies) or a pair of distinct element IDs (for relational inconsistencies).

\subsubsection{Self-Evaluation Loop}
We follow a generator-evaluator loop that refines proposals through iterative self-assessment. A simplified Python snippet of the loop function is provided below:

\begin{lstlisting}
def loop(client, image_dir, frame_id, task: str, evaluator_prompt: str, generator_prompt: str) -> tuple[str, list[dict]]:
    """Keep generating and evaluating until requirements are met."""
    memory = []
    chain_of_thought = []
    
    thoughts, result = generate(client, image_dir, frame_id, generator_prompt, task)
    memory.append(result)
    chain_of_thought.append({"thoughts": thoughts, "result": result})
    
    loop_count = 1
    while True:
        all_pass = True
        evaluation, feedback = evaluate(client, image_dir, frame_id, evaluator_prompt, result, task)
        for eval_line in evaluation.split("\n"):
            if eval_line.strip() != "PASS":
                all_pass = False
                break
        if all_pass or loop_count == 2:
            return result, evaluation
            
        context = "\n".join([
            "Previous attempts:",
            *[f"- {m}" for m in memory],
            f"\nFeedback: {feedback}"
        ])
        thoughts, result = generate(client, image_dir, frame_id, generator_prompt, task, context)
        memory.append(result)
        chain_of_thought.append({"thoughts": thoughts, "result": result})
        loop_count += 1
\end{lstlisting}

In this loop, the generator produces proposals which are then evaluated against the following criteria (as specified in the evaluator prompt):
\begin{itemize}
    \item \textbf{Category Compliance:} The edit must match the intended error category.
    \item \textbf{Atomic Modification:} Exactly one inconsistency should be introduced.
    \item \textbf{Visual Consistency:} The modified screenshot must visibly reflect the error without relying on external context.
    \item \textbf{Element Validity:} The referenced element IDs must exist in the artifact.
\end{itemize}

Only proposals receiving a "PASS" in the evaluation are retained. The loop iterates until either all criteria are met or a maximum of two iterations is reached.

\subsubsection{Prompt details for generator-evaluator proposal generation framework}

This is the task prompt as input to the o1 generator model.

\begin{lstlisting}
    task_prompt = f"""
<user input>
Your task is to modify a {category_str} to create inconsistency. For each given category of inconsistency, you will propose a modification action that introduces the inconsistency in the modified {category_str}.
    
Here's the information you'll have:
Screenshot of the urrent {category_str}: This is a screenshot of the {category_str}, with each editable element assigned a unique numerical id. Each bounding box and its respective id share the same color.
The Observation, which lists the IDs of all editable elements on the current {category_str} with their content, in the format [id] [tagType] [content], separated by "\n". Each id is mapped with the id in the screenshot. tagType is the type of the element, such as button, link, or textbox. For example, "[21] [SPAN] [Add to Wish List]" means that there is a span with id 21 and text content 'Add to Wish List' on the current {category_str}. "[23] [IMG] [Image, description: a beige powder on a white background, url: http://localhost:7770/media/catalog/product/cache/829a59e57f886f8cf0598ffca4f8a940/B/0/B074DBMG66.0.jpg]" means that there is an image on the current screen with id 23, with a description of the image and its url specified.

Here are the categories of errors you can introduce:
A. Factual Contradiction -  Direct conflict between two or more elements (text-text, text-image, or image-image). For example, The product title says "Caffeinated," while the description states "Caffeine-free." Another example: The image shows a green tea bag, but the text describes a "fruit infusion."
B. Identity Misattribution - Mislabeling of entities (objects, locations, brands, people) that conflict with other elements. Example: Product "Country of Origin: China" contradicts manufacturer info "Elmwood Inn (USA)."
C. Contextual Mismatch - Tonal, thematic, or situational incompatibility between elements. Example: A celebratory image of diplomats shaking hands paired with an article about violent clashes.
D. Quantitative Discrepancy - Numerical or statistical inconsistencies between elements. Example: A graph labeled "50%\ growth" shows flat bars.
E. Temporal/Spatial Incoherence - Implied timelines, dates, or spatial relationships that are impossible or conflicting. Example: A map labeled "North America" depicts landmarks from Europe

Here are the rules for the modification action:
The modification action you can propose to introduce inconsistency must be in the format of "Modify [id] [original_content] [new_content]": This action proposes to edit the orignal field assigned with the id to the new content to introduce inconsistency. If you propose to modify an image field, the [original_content] field should include the full content from observation including the url; the [new_content] field should be a caption describing the updated image, starting with "Image, description: ", no url needed. If you propose to modify a text field, the new content string should be about the same length as the original text field. For each inconsistency category, you should try to propose a modification action that introduces an inconsistency in that category. If you can't find a way to introduce an inconsistency in a category, you can skip it. Prioritize proposing edits on text fields over image fields.

Generate the response in the correct format. For each inconsistency, the format should be:
<proposal>
  <cat>[A-E]</cat> <-- Category letter
  <ele>[ID1,ID2]</ele> <-- Conflicting element IDs
  <mod>Modify [ID] [Original Content] [New Content]</mod> <-- Modification plan
  <rationale>Visible conflict explanation</rationale> <-- Visual verification
</proposal>
</user input>
"""
\end{lstlisting}

These are prompts for the generator and evaluator model.

\begin{lstlisting}
    evaluator_prompt = """
Evaluate the following proposals one by one for:
1. Category Compliance: Introduced inconsistency matches the category definition (A-E)
2. Atomic Modification: Introduce EXACTLY ONE inconsistency without side effects
3. Visual Consistency: Conflict visible in the modified screenshot (with NO reliance on original page knowledge or external context)
4. Element Validity: Conflict IDs exist in observations

You should be evaluating only and not attemping to solve the task.
For each proposal, only output "PASS" if all criteria are met and you have no further suggestions for improvements.
Output your evaluation concisely in the following format.

<evaluation>
PASS, NEEDS_IMPROVEMENT, or FAIL <-- For each proposal
</evaluation>
<feedback>
What needs improvement and why. <-- For proposals that need improvement
</feedback>
"""

    generator_prompt = """
Your goal is to complete the task based on <user input>. If there are feedback 
from your previous generations, you should reflect on them to improve proposals that NEEDS_IMPROVEMENT or FAIL. Leave the PASS proposals as they are. 

Output your answer concisely in the following format: 

<thoughts>
[Your understanding of the task and feedback and how you plan to improve]
</thoughts>

<response>
[Your response here]
</response>
"""
\end{lstlisting}

\subsection{Auto-Verification and Editing Process}
\label{appendix:sec:verifier and editor}

Following proposal generation, an auto-verification step filters the proposals based on format and backend constraints. Specifically:

\begin{itemize}
    \item \textbf{Edit Format Verification:} The system uses a regular expression to ensure that each proposed edit adheres to the required format: "Modify [id] [old\_content] [new\_content]".

    \item \textbf{Element Matching:} For web-sourced artifacts, the proposal’s element ID is used to locate the corresponding element and its bounding box in the metadata. The system checks that both the content and bounding box match an editable element in the HTML/PPTX structure. For image edits, the new content (a caption) is cross-referenced against an MSCOCO image database to verify its appropriateness.

    Proposals that pass these checks are automatically saved for further processing.
    
\end{itemize}

For web pages, we use the CDP to perform edit:

\begin{lstlisting}
# text edit
client.send(
    "Runtime.callFunctionOn",
    {
        "objectId": object_id, 
        "functionDeclaration": f"function() {{ this.nodeValue = '{new_content}'; }}",
        "arguments": [],
        "returnByValue": True
    }
)
# image edit
with open(new_content, "rb") as image_file:
    img = Image.open(image_file)
    new_image_width, new_image_height = img.size  # get original width and height for resizing
    aspect_ratio = new_image_width / new_image_height
    if w / h > aspect_ratio:
        w, h = w, int(w / aspect_ratio)
    else:
        w, h = int(h * aspect_ratio), h
    img = img.resize((w, h), Image.Resampling.LANCZOS)
    buffer = BytesIO()
    img.save(buffer, format="JPEG")
    buffer.seek(0)
    base64_image = base64.b64encode(buffer.read()).decode("utf-8")
    new_image = f"data:image/jpeg;base64,{base64_image}"
client.send(
    "Runtime.callFunctionOn",
    {
        "objectId": object_id,
        "functionDeclaration": f"""
            function() {{
                this.src = '{new_image}';
            }}
            """,
        "arguments": [],
        "returnByValue": True
    }
)
\end{lstlisting}

For Zenodo presentation, we use the \texttt{python-pptx} library:

\begin{lstlisting}
# text edit
if target_shape.has_text_frame: # text edit
    text_frame = target_shape.text_frame
    for paragraph in text_frame.paragraphs:
        for run in paragraph.runs:
            if edit_info["old_content"] in run.text:
                try:
                    run.text = run.text.replace(edit_info["old_content"], edit_info["new_content"])
                    success = True
                    break
                except:
                    success = False
# image edit
left, top, orig_width, orig_height = target_shape.left, target_shape.top, target_shape.width, target_shape.height
pic = target_shape._element
pic.getparent().remove(pic)
new_image_path = f"{coco_image_dir}/{edit_info['new_img_path']}"
with Image.open(new_image_path) as img:
    new_width, new_height = img.size
new_aspect = new_width / new_height
orig_aspect = orig_width / orig_height
if new_aspect > orig_aspect:
    scaled_width = orig_width
    scaled_height = int(scaled_width / new_aspect)
else:
    scaled_height = orig_height
    scaled_width = int(scaled_height * new_aspect)
new_left = left + (orig_width - scaled_width) // 2
new_top = top + (orig_height - scaled_height) // 2
try:
    slide.shapes.add_picture(  # Add the new image in the same location and size
        new_image_path, new_left, new_top, scaled_width, scaled_height
    )
    success = True
except:
    success = False
\end{lstlisting}

\section{Qualitative Example and Analysis}
\label{appendix:sec:example_main settings}

\begin{figure*}[h!]
\setlength\tabcolsep{0pt}
\setlength{\abovecaptionskip}{0.1cm}
    \centering
    \includegraphics[width=\linewidth]{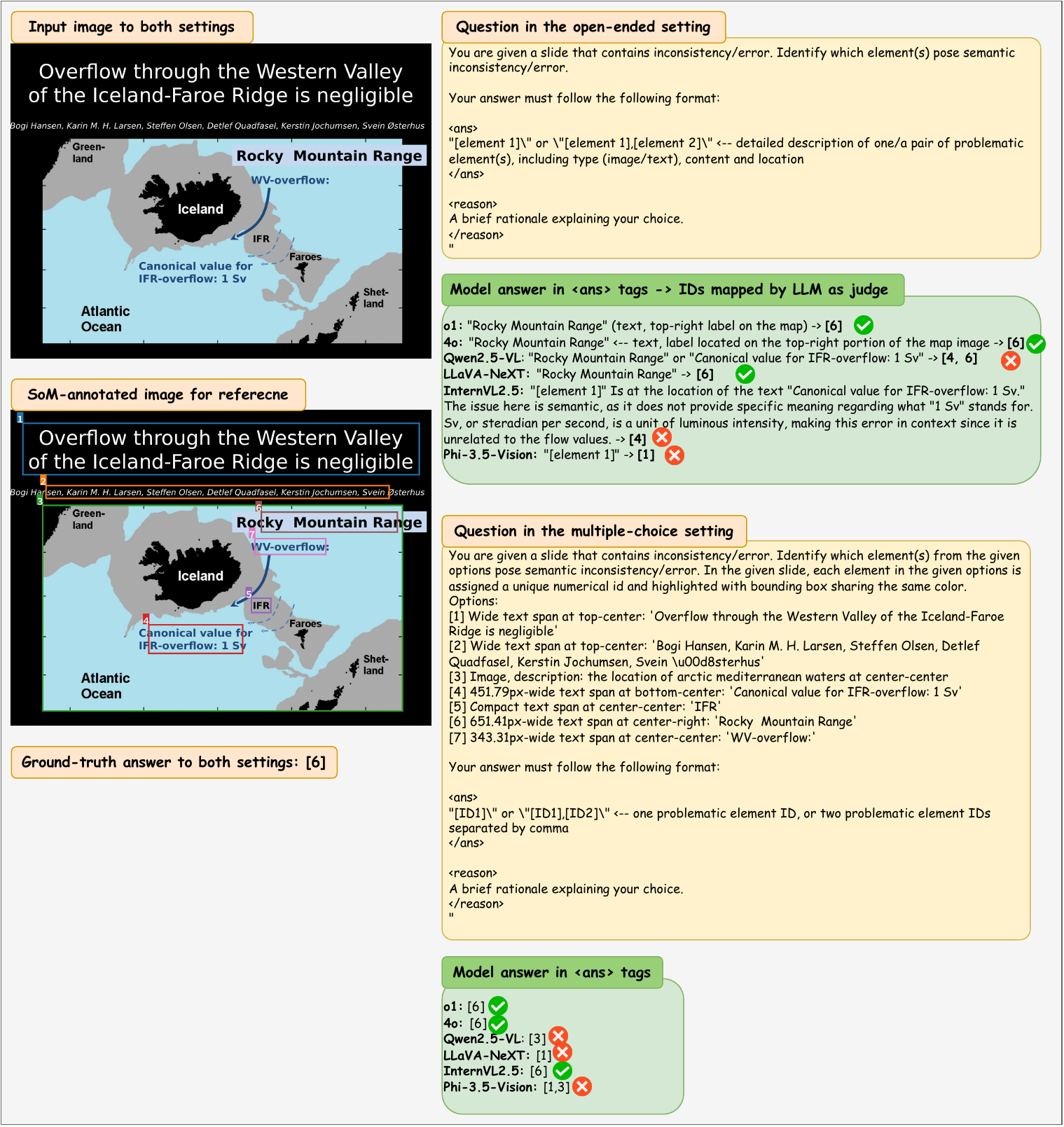}
    \caption{A test sample with model responses under the two main settings in MMIR: open-ended and multiple-choice.}
    \vspace{-10pt}
    \label{fig:qualitative_example}
\end{figure*}

Figure~\ref{fig:qualitative_example} illustrates a test sample with model responses under the two main settings in MMIR: open-ended and multiple-choice.

Beyond the single illustration, we performed a qualitative analysis of the top-performing models’ (o1, GPT-4o) performance in the open-ended format, examining an additional 50 randomly selected samples. Our analysis revealed multiple common failure modes:

\begin{itemize}
    \item \textbf{Contextual misunderstanding}: The model frequently flagged false positives when required to integrate contextual information, such as distinguishing between user-selected and available options. For instance, on a shopping website listing a “Bath Body Brush,” with a photo showing a pink brush and selectable color options (Pink, Green, etc.), the model incorrectly flagged inconsistency due to confusion over color selection context (the photo and the unselected option of “Green”), despite clear visual alignment between the image and selected color option “Pink”.
    \item \textbf{Indirect semantic reasoning}: Errors often arose when inconsistencies required inference beyond direct semantic contradictions. The model struggled notably with subtle contradictions, such as mismatches between implied and explicit information (e.g., temporal contexts like "limited-time offers" vs. static pricing information).
    \item \textbf{Complex textual layouts}: The model's accuracy significantly decreased when inconsistencies were embedded within lengthy or dense textual sections, such as detailed paragraphs in product descriptions, compared to simpler fields like product titles.
\end{itemize}

\section{Model Application Details}
\label{appendix:sec:model_detail}

Here are the generation methods for the open-sourced models.

For \textbf{o1} and \textbf{GPT-4o}, we utilized the API following API guidelines available at \url{https://platform.openai.com/docs/models#gpt-4o}. 

For \textbf{Qwen2.5-VL}, we implemented the 72B and 7B versions following the official repository: \url{https://github.com/QwenLM/Qwen2.5-VL}. 

For \textbf{Llama-3.2}, we utilized the API following the API guidelines available at \url{https://openrouter.ai/qwen/qwen-2.5-72b-instruct}.

For \textbf{LLaVA-NeXT}, we followed the implementation from \url{https://github.com/LLaVA-VL/LLaVA-NeXT}. 

For \textbf{InternVL2.5} we implemented the 8B version at \url{https://github.com/OpenGVLab/InternVL}.

For \textbf{Phi-3.5-Vision} we implemented the 4B version at \url{https://github.com/instill-ai/models/tree/main/phi-3-5-vision}.

\section{Data Release}

We will publicly release a comprehensive dataset that includes the artifacts and question-answer pairs in both the open-ended and multiple-choice settings. The licensing terms for the artifacts will follow those set by the respective dataset creators, as referenced in this work, while the curated artifacts will be provided under the MIT License. 
Additionally, our release will include standardized evaluation protocols, and evaluation scripts to facilitate rigorous assessment. The entire project will be open-sourced, ensuring free access for research and academic purposes.

\end{document}